\documentclass{article}

\usepackage[final]{neurips_data_2024}

\usepackage[utf8]{inputenc} 
\usepackage[T1]{fontenc}    
\usepackage{hyperref}       
\usepackage{url}            
\usepackage{booktabs}       
\usepackage{amsfonts}       
\usepackage{nicefrac}       
\usepackage{microtype}      
\usepackage{xcolor}         
\usepackage{listings}

\definecolor{codegreen}{rgb}{0,0.6,0}
\definecolor{codegray}{rgb}{0.5,0.5,0.5}
\definecolor{codepurple}{rgb}{0.58,0,0.82}
\definecolor{backcolour}{rgb}{0.95,0.95,0.92}

\lstdefinestyle{prettypython}{
    backgroundcolor=\color{backcolour},   
    commentstyle=\color{codegreen},
    keywordstyle=\color{magenta},
    numberstyle=\tiny\color{codegray},
    stringstyle=\color{codepurple},
    basicstyle=\ttfamily\footnotesize,
    breakatwhitespace=false,         
    breaklines=true,                 
    captionpos=b,                    
    keepspaces=true,                 
    numbers=left,                    
    numbersep=5pt,                  
    showspaces=false,                
    showstringspaces=false,
    showtabs=false,                  
    tabsize=2
}

\usepackage[capitalize,noabbrev]{cleveref}

\usepackage{graphicx}
\usepackage{subfigure}

\newcommand{\rebuttal}[1]{\textcolor{black}{#1}}

\hypersetup{
  colorlinks   = true, 
  urlcolor     = blue, 
  linkcolor    = blue, 
  citecolor   = orange 
}

\title{CALE: Continuous Arcade Learning Environment}

\author{%
  Jesse Farebrother \\
  McGill University\\
  Mila - Québec AI Institute\\
  Google DeepMind \\
  \texttt{jfarebro@cs.mcgill.ca} \\
  \And
  Pablo Samuel Castro \\
  Google DeepMind \\
  Université de Montréal \\
  Mila - Québec AI Institute\\
  \texttt{psc@google.com} \\
}

\begin{document}

\maketitle

\begin{abstract}
  We introduce the Continuous Arcade Learning Environment (CALE), an extension of the well-known Arcade Learning Environment (ALE) \citep{bellemare13ale}. The CALE uses the same underlying emulator of the Atari 2600 gaming system (Stella), but adds support for continuous actions. This enables the benchmarking and evaluation of continuous-control agents (such as PPO \citep{schulman17ppo} and SAC \citep{haarnoja18sac}) and value-based agents (such as DQN \citep{mnih2015humanlevel} and Rainbow \citep{hessel18rainbow}) on the same environment suite. We provide a series of open questions and research directions that CALE enables, as well as initial baseline results using Soft Actor-Critic. CALE is available \rebuttal{as part of the ALE at \url{https://github.com/Farama-Foundation/Arcade-Learning-Environment}}.
\end{abstract}

\section{Introduction}
\label{sec:intro}
Generally capable autonomous agents have been a principal objective of machine learning research, and in particular reinforcement learning, for many decades. {\em General} in the sense that they can handle a variety of challenges; {\em capable} in that they are able to ``solve'' or perform well on these challenges; and they are able to learn {\em autonomously} by interacting with the system or problem by exercising their {\em agency} (e.g. making their own decisions). While deploying and testing on real systems is the ultimate goal, researchers usually rely on academic benchmarks to showcase their proposed methods. It is thus crucial for academic benchmarks to be able to test generality, capability, and autonomy.
 
\citet{bellemare13ale} introduced the Arcade Learning Environment (ALE) as one such benchmark. The ALE is a collection of challenging and diverse Atari 2600 games where agents learn by directly playing the games; as input, agents receive a high dimensional observation (the ``pixels'' on the screen), and as output they select from one of 18 possible actions (see \cref{sec:VCStoALE}). While some research had already been conducted on a few isolated Atari 2600 games \citep{cobo11automatic,hausknecht12hyperneat,bellemare21investigating}, the ALE's significance was to provide a unified platform for research and evaluation across more than $100$ games. Using the ALE, \citet{mnih2015humanlevel} demonstrated, for the first time, that reinforcement learning (RL) combined with deep neural networks could play challenging Atari 2600 games with super-human performance. Much like how ImageNet \citep{deng2009imagenet} ushered in the era of Deep Learning \citep{lecun15deep}, the Arcade Learning Environment spawned the advent of Deep Reinforcement Learning.

In addition to becoming one of the most popular benchmarks for evaluating RL agents, the ALE has also evolved with new extensions, including stochastic transitions \citep{machado18revisiting}, various game modes and difficulties \citep{machado18revisiting,farebrother18generalization}, and multi-player support \citep{terry2020arcade}. What has remained constant is the suitability of this benchmark for testing {\em generality} (there is a wide diversity of games), {\em capability} (many games still prove challenging for most modern agents), and {\em agency} (learning typically occurs via playing the game).

There are a number of design choices that have become standard when evaluating agents on the ALE, and which affect the overall learning dynamics. These choices involve modifying the temporal dynamics through frame skipping; adjusting the input observations with frame stacking, grey-scaling, and down-sampling; and converting the range of joystick movements into a standardized set of 18 discrete actions to be shared across all games.\footnote{Certain games, such as Pong and Breakout, were originally played using a different set of paddle controllers, but were given the same action space in the ALE.} The design of the action space resulted in a rather profound impact on the type of research conducted on the ALE.
In particular, {\em it is only compatible with discrete-action agents}. This has led to certain classes of agents, often based on Q-learning \citep{watkins89thesis}, \rebuttal{to focus primarily} on the ALE\rebuttal{. On the other hand, agents} based on policy gradient \citep{sutton99pg} or actor-critic \citep{konda99ac} methods, \rebuttal{while sometimes evaluating on the ALE by using discrete variants, tend to focus} on entirely different benchmarks, such as MuJoCo \citep{todorov19mujoco} or DM-Control \citep{Tassa2018DeepMindCS}.

In this paper, we introduce the Continuous Arcade Learning Environment (CALE) that introduces a continuous action space making for an interface that more closely resembles how humans interact with the Atari 2600 console.
Our work enables the evaluation of both discrete and continuous action agents on a single unified benchmark, providing a unique opportunity to gain an understanding of the challenges associated with the action-space of the agent.
Additionally, we present baselines with the popular Soft-Actor Critic \citep[SAC;][]{haarnoja18sac} algorithm that underscore the need for further research towards general agents capable of handling diverse domains.
Finally, we identify key challenges in representation learning, exploration, transfer, and offline RL, paving the way for more comprehensive research and advancements in these areas.

\begin{figure}[!t]
    \centering
    \raisebox{0.3\height}{\includegraphics[width=0.2\textwidth]{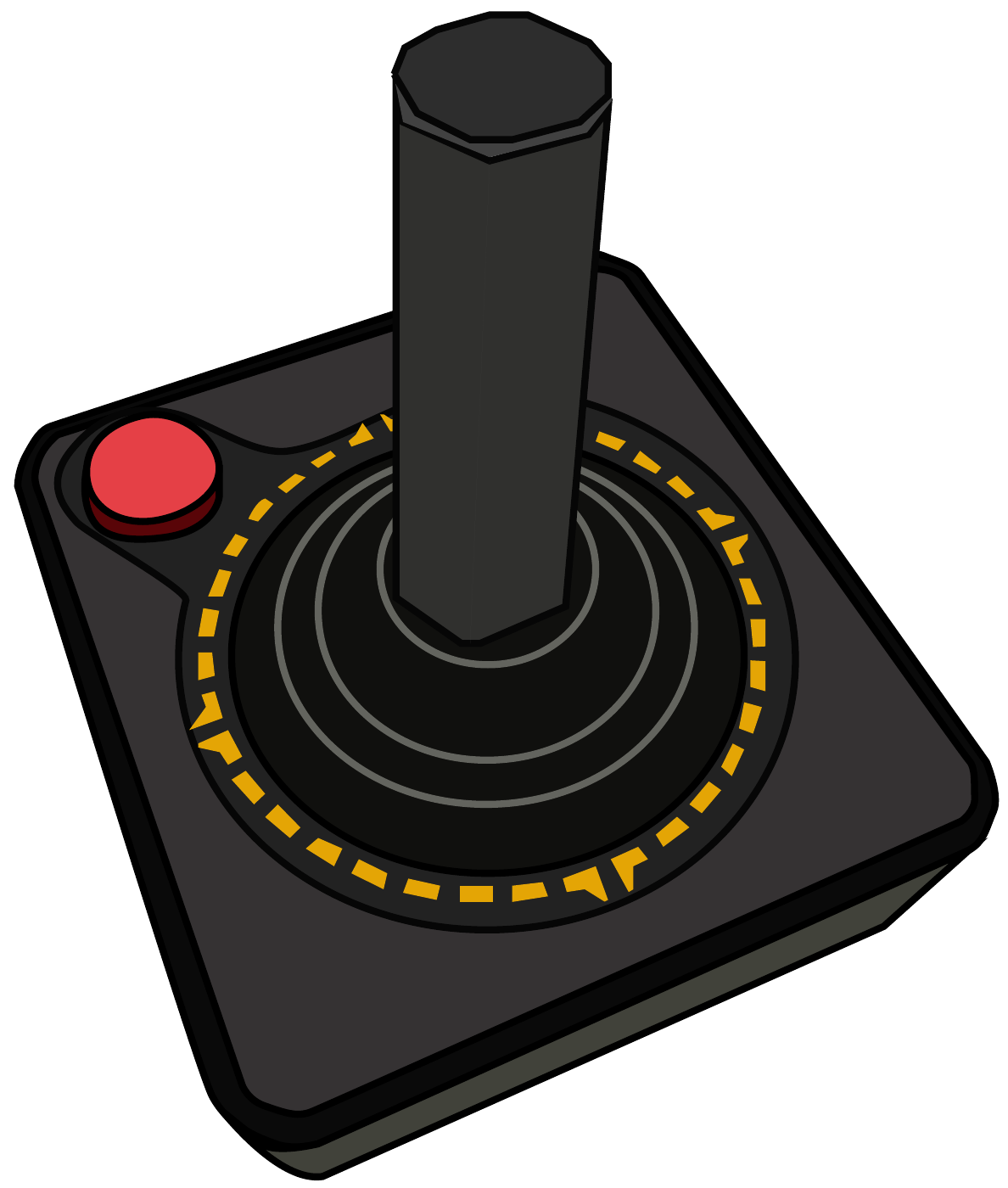}}%
    \includegraphics[width=0.6\textwidth]{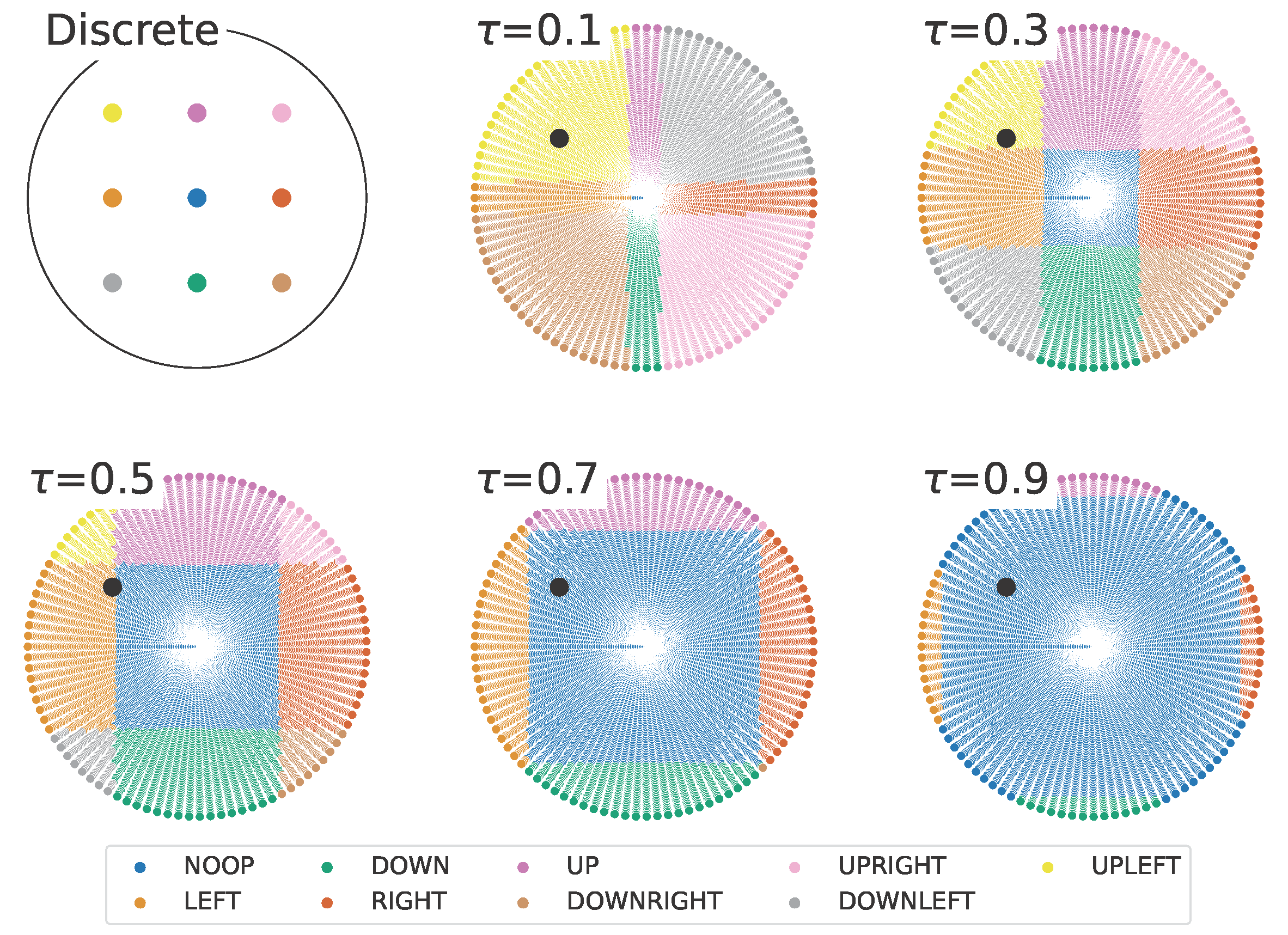}
    \caption{{\bf Left panel:} Atari CX10 controller. {\bf Right panel:} Discrete joystick positions (top left) versus continuous joystick positions with varying values of the threshold $\tau$. The black circle corresponds to a joystick at position $(r,\theta)=(0.61, 2.53)$. 
    }
    \label{fig:actionMappings}
\end{figure}

\section{From Atari VCS to the Arcade Learning Environment}
\label{sec:VCStoALE}

The Atari Video Computer System (VCS), later renamed the Atari 2600, is a pioneering gaming console developed in the late 1970s that aimed to bring the arcade experience to the home.
Game designers had to operate under a variety of constraints, including writing code that could execute in time with the electron beam displaying graphics on the CRT screen and rendering graphics using the limited set of primitives provided by the system. Although designed to support a variety of controllers, the majority of games were played with an Atari CX10 ``digital'' controller (see left panel of \cref{fig:actionMappings}). Players move a joystick along two axes to trigger one of nine discrete events (corresponding to three positions on each axis) on the Atari VCS. Combined with a ``fire'' button, this results in 18 possible events the user could trigger.\footnote{For the interested reader, \citet{montfort09racing} provide a great historical overview of the design and development of the Atari VCS.}

The Atari 2600 was one of the first widely popular home gaming devices and even became synonymous with ``video games'', marking the beginning of exponential growth in the video game industry over the following decades. A likely reason for its popularity was the use of external cartridges containing read-only memory (ROM), which allowed for a scalable plug and play experience. Over 500 games were developed for the original console, offering a wide variety of game dynamics and challenges that appealed to an ever-growing audience. 
As personal computing became more widespread, emulators such as Stella \citep{stella96} emerged, allowing enthusiasts to continue playing Atari 2600 games without needing the original hardware.

Building upon the Stella emulator, \citet{bellemare13ale} introduced the Arcade Learning Environment (ALE) as a challenging and diverse environment suite for evaluating generally capable agents. The authors argue the ALE contains three crucial features which render it a meaningful baseline for agent evaluation: {\bf variety} -- it contains a diverse set of games; {\bf relevance} -- the varied challenges presented are reflective of challenges agents may face in practically-relevant environments; and {\bf independence} -- it was developed independently for human enjoyment, free from researcher bias.

This seminal benchmark was used by \citet{mnih2015humanlevel} to showcase super-human performance when combining temporal-difference learning \citep{sutton84thesis} with deep neural networks. The performance of their DQN agent was compared against the average performance of a single human expert; these average human scores now serve as the standard way to normalize and aggregate scores on the ALE \citep{agarwal21deep}. Since its introduction, numerous works have improved on DQN, such as Double DQN \citep{hasselt16doubleDQN}, Rainbow \citep{hessel18rainbow}, C51 \citep{bellemare17distributional}, A3C 
\citep{mnih16asynchronous}, IMPALA \citep{espeholt18impala}, R2D2 \citep{kapturowski2018recurrent}, and Agent57 \citep{badia20agent57};  the ALE continues to serve as a simulator-based test-bed for evaluating new algorithms and conducting empirical analyses, especially with limited compute budgets.

\section{CALE: Continuous Arcade Learning Environment}
\label{sec:cale}

The original Atari CX10 controller (left panel of \cref{fig:actionMappings}) used a series of pins to signal to the processor when the joystick is in one of nine distinct positions, visualized in the `Discrete' sub-panel in \cref{fig:actionMappings} \citep{sivakumaran86electronic}. When combined with a boolean ``fire'' button, this results in 18 distinct joystick {\em events}. Indeed, player control in the Stella emulator is built on precisely these distinct events \citep{stella96}, and they also correspond to the 18 actions chosen by the ALE.

However, although the resulting events are discrete, the range of joystick motion available to players is continuous. We add this capability by introducing the {\bf C}ontinuous {\bf A}rcade {\bf L}earning {\bf E}nvironment (CALE), which switches from a set of 18 discrete actions to a three-dimensional continuous action space. Specifically, we use the first two dimensions to specify the polar coordinates $(r, \theta)$ in the unit circle corresponding to all possible joystick positions, while the last dimension is used to simulate pressing the ``fire'' button. Concretely, the action space is $[0, 1]\times [-\pi, \pi] \times [0, 1]$.
 The implementation of CALE is available \rebuttal{as part of the ALE at  \url{https://github.com/Farama-Foundation/Arcade-Learning-Environment}} (under GPL-2.0 license). See \cref{sec:howto} for usage instructions.

As in the original CX10 controller, this continuous action space still needs to trigger discrete events. For this, we use a threshold $\tau$ to demarcate the nine possible position events the joystick can trigger. \cref{fig:actionMappings} illustrates these for varying values of $\tau$, as well as the different events triggered when the joystick is at position $(r,\theta)=(0.61, 2.53)$. As can be seen, lower values of $\tau$ result in more sensitive control, while higher values can result in less responsive controllers, even to the point of completely occluding certain events (the corner events are unavailable when $\tau=0.9$, for example).

It is worth highlighting that, since CALE is essentially a wrapper around the original ALE, it is {\em only} changing the agent action space. Since both discrete and continuous actions ultimately trigger the same events, the underlying game mechanics and learning environment remain unchanged. This is an important point, as it means that we now have a {\em unified} benchmark on which to directly compare discrete and continuous control agents.

An important difference is that the ALE supports ``minimal action sets'', which reduce the set of available actions from 18 to the minimum required to play the game. For example, in Breakout only the LEFT, RIGHT, and FIRE events have an effect on game play, resulting in a minimal set of 4 actions. By default, minimum action sets are enabled in the ALE and used by many existing implementations \citep{castro18dopamine}. Given the manner in which continuous actions have been parameterized, this minimal action set is unavailable when running with the CALE. Thus, for many games, continuous-action agents trained on the CALE may be at a disadvantage when compared with discrete-action agents trained on the ALE (see comparison in \cref{sec:comparisonToDiscrete} and \cref{fig:actionDistributions} in particular). For completeness, we list the minimum action sets for all 60 games in \cref{sec:aleGameSpecs}.

\section{Baseline results}
\label{sec:baseline}
We present a series of baseline results on CALE using the soft actor-critic agent \citep[SAC;][]{haarnoja18sac}. SAC is an off-policy continuous-control method that modifies the standard Bellman backup with entropy maximization \citep{ziebart08maximum,ziebart10phd}. DQN and the agents derived from it are also off-policy methods, thereby rendering SAC a more natural choice for this initial set of baselines than other continuous control methods such as PPO. We use the SAC implementation and experimental framework provided by Dopamine \citep{castro18dopamine}. We detail our experimental setup and hyper-parameter selection below, and provide further details in \cref{sec:sacHypers}.

\subsection{Experimental setup}
We use the evaluation protocol proposed by \citet{machado18revisiting}. Namely, agents are trained for 200 million frames with ``sticky actions'' enabled, 4 stacked frames, a frame-skip of 4, and on 60 games. Additionally, we use the Atari 100k benchmark introduced by \citet{kaiser2020model}, which evaluates agents using only 100,000 agent interactions (corresponding to 400,000 environment steps due to frame-skips) over 26 games. The Atari 100k benchmark has become a popular choice for evaluating the sample efficiency of RL agents \citep{doro2023sampleefficient,schwarzer23bbf}. We follow the evaluation protocols of \citet{agarwal21deep} and report aggregate results using interquartile mean (IQM), with shaded areas representing 95\% stratified bootstrap confidence intervals. All experiments were run on P100 GPUs; the 200M experiments took between 5-7 days to complete training, while the 100K experiments took between 1 and 2 hours to complete.

\subsection{Threshold selection}
\begin{figure}[!t]
    \centering
    {\includegraphics[width=0.8\textwidth]{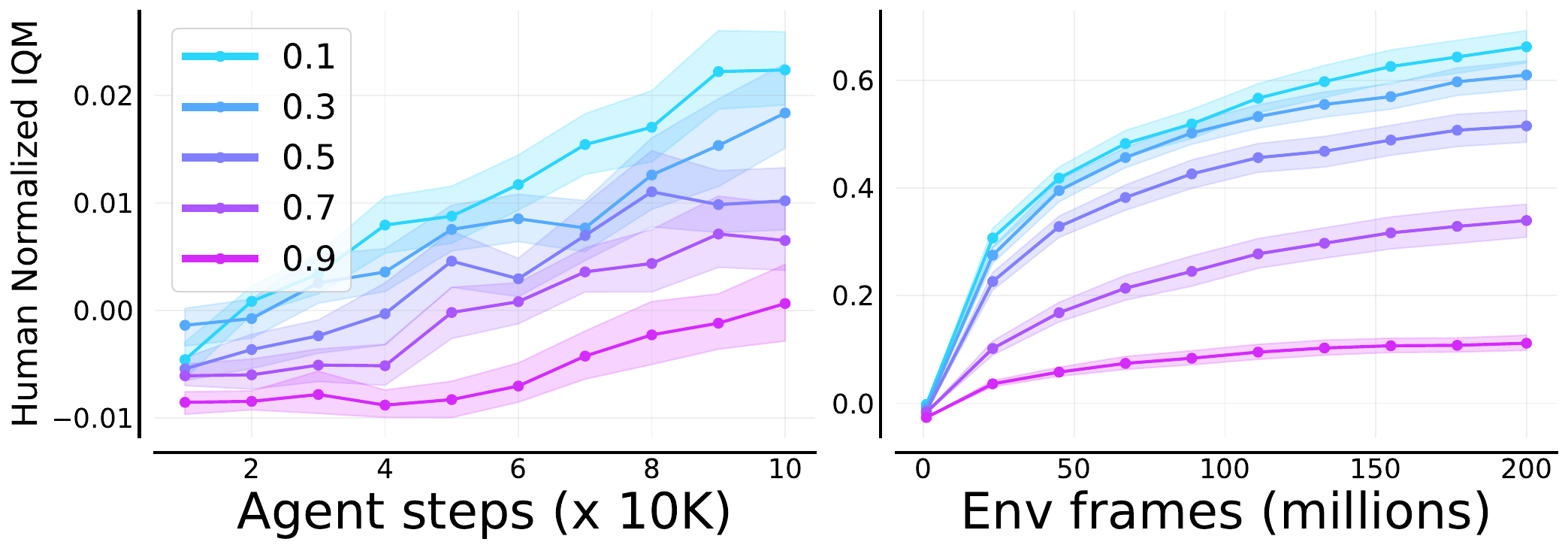}}
    \caption{\rebuttal{CALE} comparison with varying $\tau$ on the 100k (left) and 200m (right) training regimes.}
    \label{fig:thresholdComparison}
\end{figure}

As mentioned in \cref{sec:cale}, the choice of threshold $\tau$ affects the overall performance of the agents. Consistent with intuition, \cref{fig:thresholdComparison} demonstrates that higher values of $\tau$ result in degraded performance. For the remaining experimental evaluations we set $\tau$ to $0.5$. This choice has consequences for SAC, due to the way its action outputs are initialized, which we discuss in the next subsection.

\subsection{Network architectures}
\label{sec:netArchs}
Given an input state $x\in\mathcal{X}$, the neural networks used by actor-critic methods usually consist of an ``encoder'' $\phi:\mathcal{X}\rightarrow\mathbb{R}^d$, and actor and critic heads $\psi_{A}:\mathbb{R}^d\rightarrow \mathcal{A}$ and $\psi_{C}:\mathbb{R}^d\rightarrow \mathbb{R}$, respectively, where $\mathcal{A}$ is the (continuous) action space. Typically the action outputs are assumed to be Gaussian distributions with mean $\mu$ and standard deviation $\sigma$. Thus, for a state $x$, the {\em value} of the state is $\psi_{C}(\phi(x))$ and the action selected is distributed according to $\psi_{A}(\phi(x))$.

The SAC implementation we use initializes $\mu$ in the middle of the action ranges. Thus, for the CALE action space, $\mu$ is initialized at $(0.5, 0.0, 0.5)$. With $\tau=0.5$, this means the $r$ and ``fire'' dimensions will be initially straddling the threshold, where action variations are most significant. On the other hand, this initialization produces an initial $\theta$ value of $0.0$, which results in an initial bias towards the RIGHT event (since polar coordinates $(0.5, 0.0)$ correspond to $(0.5, 0.0)$ Cartesian coordinates). See \cref{fig:actionDistributions} and the surrounding discussion for more details.

For all our experiments we use a two-layer multi-layer perceptron (MLP) with 256 hidden units each for both $\psi_{A}$ and $\psi_{C}$. \citet{haarnoja18sac} benchmarked SAC on non-pixel environments, where $\phi$ consisted purely of an MLP. For pixel-based environments like the ALE, however, convolutional networks are typically preferred encoders. \citet{yarats21improving} proposed a convolutional encoder network for SAC (based on the encoder proposed by \citet{Tassa2018DeepMindCS} for the DeepMind Control Suite), which was further used by \citet{yarats2021image}. We refer to this encoder as $\phi_{SAC}$. We refer to the three-layer convolutional encoder originally used by DQN \citep{mnih2015humanlevel} (and used by most DQN-based algorithms) as $\phi_{DQN}$.

\begin{figure}[!t]
    \centering
    {\includegraphics[width=0.8\textwidth]{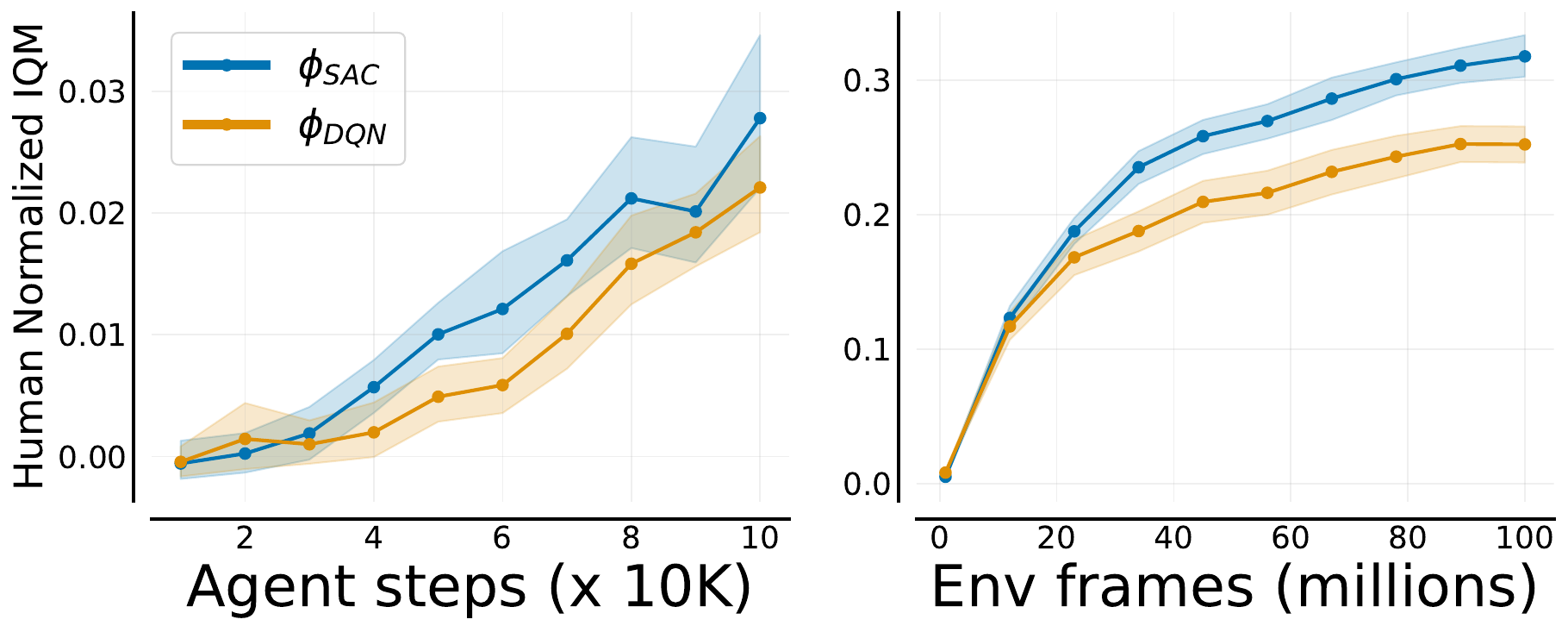}}
    \caption{\rebuttal{CALE} comparison of $\phi_{SAC}$ and $\phi_{DQN}$ on the 100k (left) and 200m (right) training regimes.}
    \label{fig:encoderComparison}
\end{figure}

As \cref{fig:encoderComparison} demonstrates, $\phi_{SAC}$ outperforms $\phi_{DQN}$ in both the 100K and 200M training regimes. Although DQN has not been explicitly tested with $\phi_{SAC}$, it begs the question of whether certain algorithms benefit from certain types of encoder architectures over others; this relates to questions of representation learning, which we discuss below.

\subsection{Exploration strategies}
\label{sec:exploration}
\begin{figure}[!h]
    \centering
    {\includegraphics[width=0.8\textwidth]{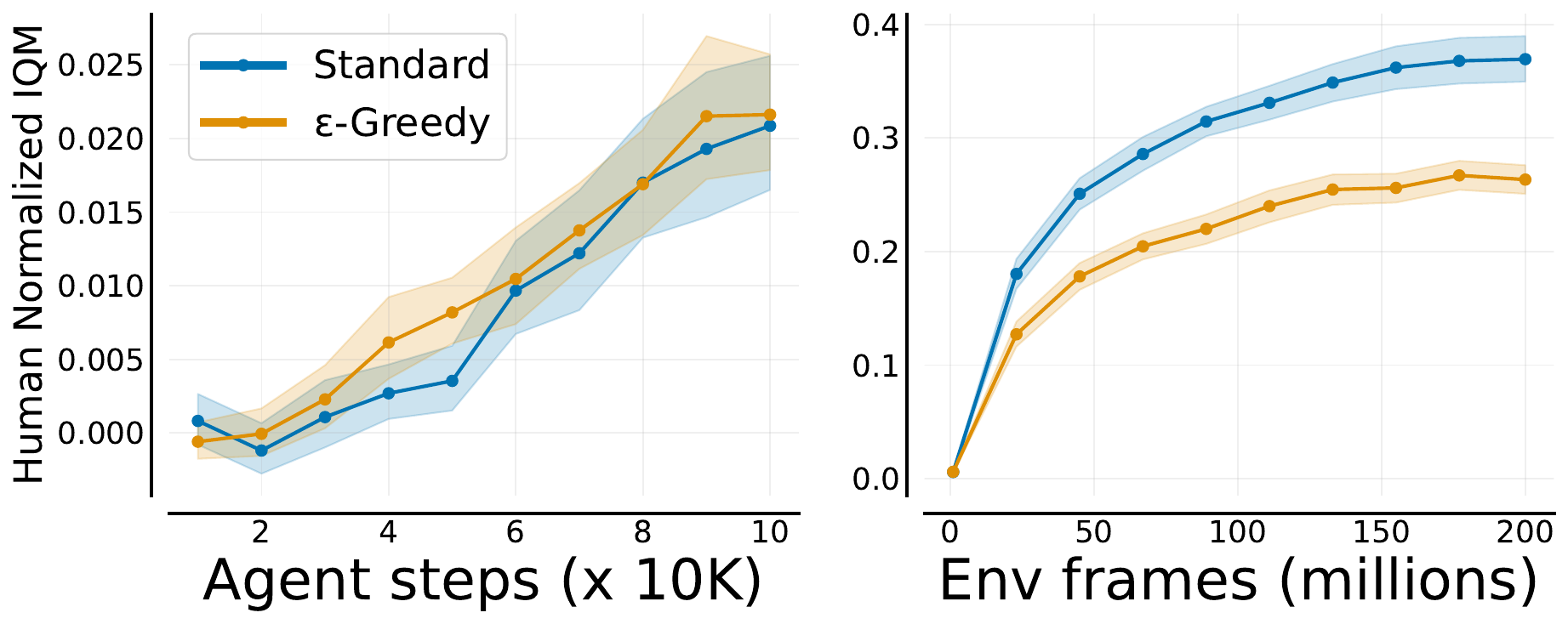}}
    \caption{\rebuttal{CALE} comparison of default SAC exploration with the more common $\epsilon$-greedy exploration used in discrete action agents on the 100k (left) and 200m (right) training regimes.}
    \label{fig:explorationComparison}
\end{figure}

Due to its objective including entropy maximization and the fact that the actor is parameterized as a Gaussian distribution, SAC induces a natural exploration strategy obtained by sampling from $\psi_{A}$ (and simply using $\mu$ when acting greedily). We refer to this as the {\bf standard} exploration strategy. However, the exploration strategy typically used on the ALE is {\bf $\epsilon$-greedy}, where actions are chosen randomly with probability $\epsilon$; a common choice for ALE experiments is to start $\epsilon$ at $1.0$ and decay it to $0.01$ over the first million environment frames. For our continuous action setup we sample uniformly randomly in $[0, 1]\times [-\pi, \pi] \times [0, 1]$ with probability $\epsilon$. Perhaps surprisingly, {\bf standard} outperforms {\bf $\epsilon$-greedy} exploration in the 200 million training regime, as demonstrated in \cref{fig:explorationComparison}. This may be due to the way the action outputs are parameterized, and merits further inquiry.

\subsection{Comparison to existing discrete-action agents}
\label{sec:comparisonToDiscrete}
\begin{figure}[!h]
    \centering
    {\includegraphics[width=0.8\textwidth]{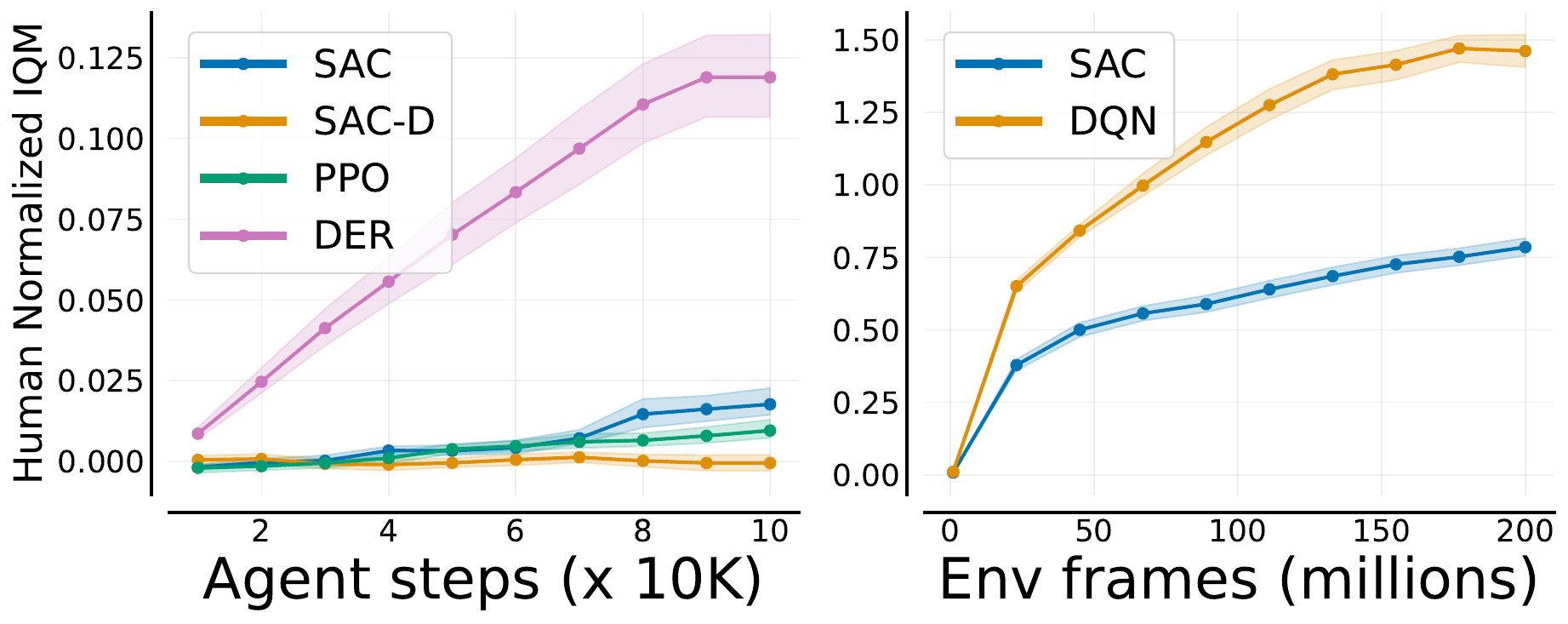}}
    \caption{Aggregate comparison of SAC \rebuttal{and PPO} on the CALE with DER \rebuttal{and SAC-D} on the ALE \citep{vanHasselt2019der} (left), and DQN on the the ALE \citep{mnih2015humanlevel} (right).}
    \label{fig:dqnAggregateComparison}
\end{figure}

We compare the performance of our SAC baseline against DQN in the 200 million training regime, given that both are off-policy methods which have similar value estimation methods; for the 100k training regime we compare against Data-Efficient Rainbow~\citep[DER;][]{vanHasselt2019der}, a popular off-policy method for this regime that is based on DQN. As \cref{fig:dqnAggregateComparison} shows, SAC dramatically under-performs, relative to both these methods. While there may be a number of reasons for this, the most likely one is the fact that SAC was not tuned for CALE, whereas both DER and DQN were tuned specifically for the ALE.

We additionally compared to a version of SAC with a categorical action parameterization which allows us to run it on the original ALE. The hyper-parameters (listed in \cref{sec:sacHypers}) are based on those suggested by \citet{christodoulou19sac}. Surprisingly, this discrete-action SAC on the ALE agent dramatically underperforms even against our continuous-action SAC on the CALE.

\begin{figure}[!h]
    \centering
    {\includegraphics[width=\textwidth]{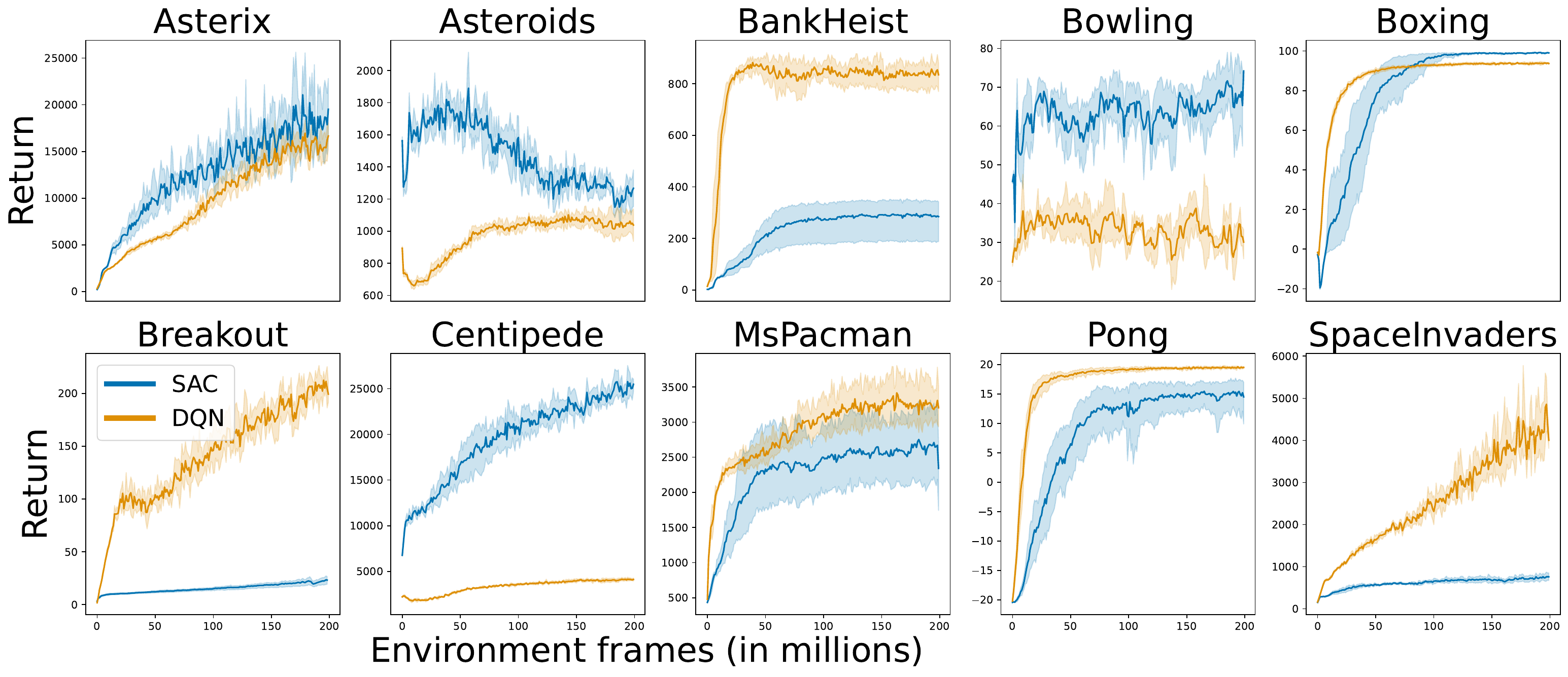}}
    \caption{\rebuttal{CALE} comparison of SAC with DQN (using the default Dopamine implementation \citep{castro18dopamine}) on a selection of games. Returns averaged over 5 independent runs, with shaded areas representing 95\% confidence intervals.}
    \label{fig:subgames200m}
\end{figure}

Aggregate performance curves can often conceal interesting per-game differences. Indeed, \cref{fig:subgames200m} demonstrates that SAC can sometimes surpass the performance of DQN (Asteroids, Bowling, Centipede), sometimes have comparable performance (Asterix, Boxing, MsPacman, Pong), and sometimes under-perform (BankHeist, Breakout, SpaceInvaders). Minimal action sets (as discussed in \cref{sec:cale}) do not appear to correlate with these performance differences (Bowling, Pong and SpaceInvaders all use a minimal set of 6 actions in the ALE); similarly, reward distributions (as we will discuss below) do not appear to correlate with performance differences between these two agents either. The differences may be due to differences in transition dynamics, as well as exploration challenges, which we discuss below.

\begin{figure}[!t]
    \centering
    {\includegraphics[width=0.9\textwidth]{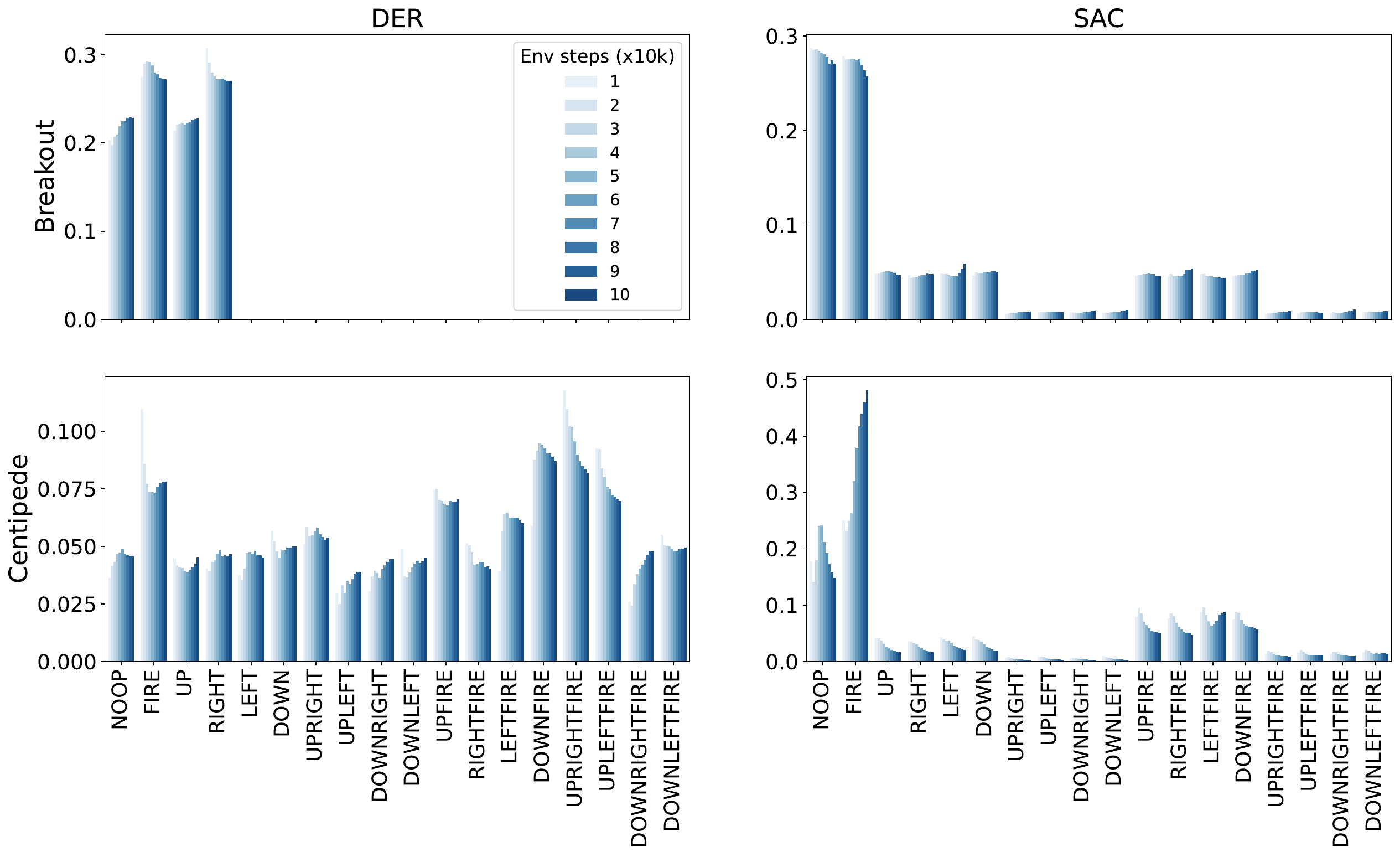}}
    \caption{Comparison of joystick event distributions during training of DER on the ALE (left column) and SAC on the CALE (right column) in the 100K benchmark. These are on a single run when training on Breakout (where DQN strongly outperforms SAC) and Centipede (where SAC strongly outperforms DQN).}
    \label{fig:actionDistributions}
\end{figure}

\cref{fig:actionDistributions} displays the distribution of discrete joystick events triggered by both DER and SAC and confirms that, while some games like Breakout on the ALE only trigger 4 events, most events are triggered on the CALE. It is interesting to observe that, as discussed in \cref{sec:netArchs}, SAC has a bias towards the RIGHT action, due to the action parameterization and initialization. 

\section{Comparison to other continuous control environments}
\label{sec:comparisonCCEnvs}
The most commonly used continuous control methods are centered around robotics tasks such as locomotion \citep{todorov19mujoco,wolczyk21continual,khazatsky2024droid}, where transition dynamics are relatively smooth and can thus be approximated reasonably well with Gaussian distributions. This assumption is often critical to certain methods, for instance in the reparameterization of the Wasserstein-2 for the DBC algorithm proposed by \citet{zhang2021learning}). \rebuttal{Thus, the ``non-smoothness'' of the CALE yields a novel challenge for continuous control agents, which could help us better understand, and improve, them.} 

\begin{figure}[!t]
    \centering
    \begin{minipage}{0.48\linewidth}
    \includegraphics[width=\textwidth]{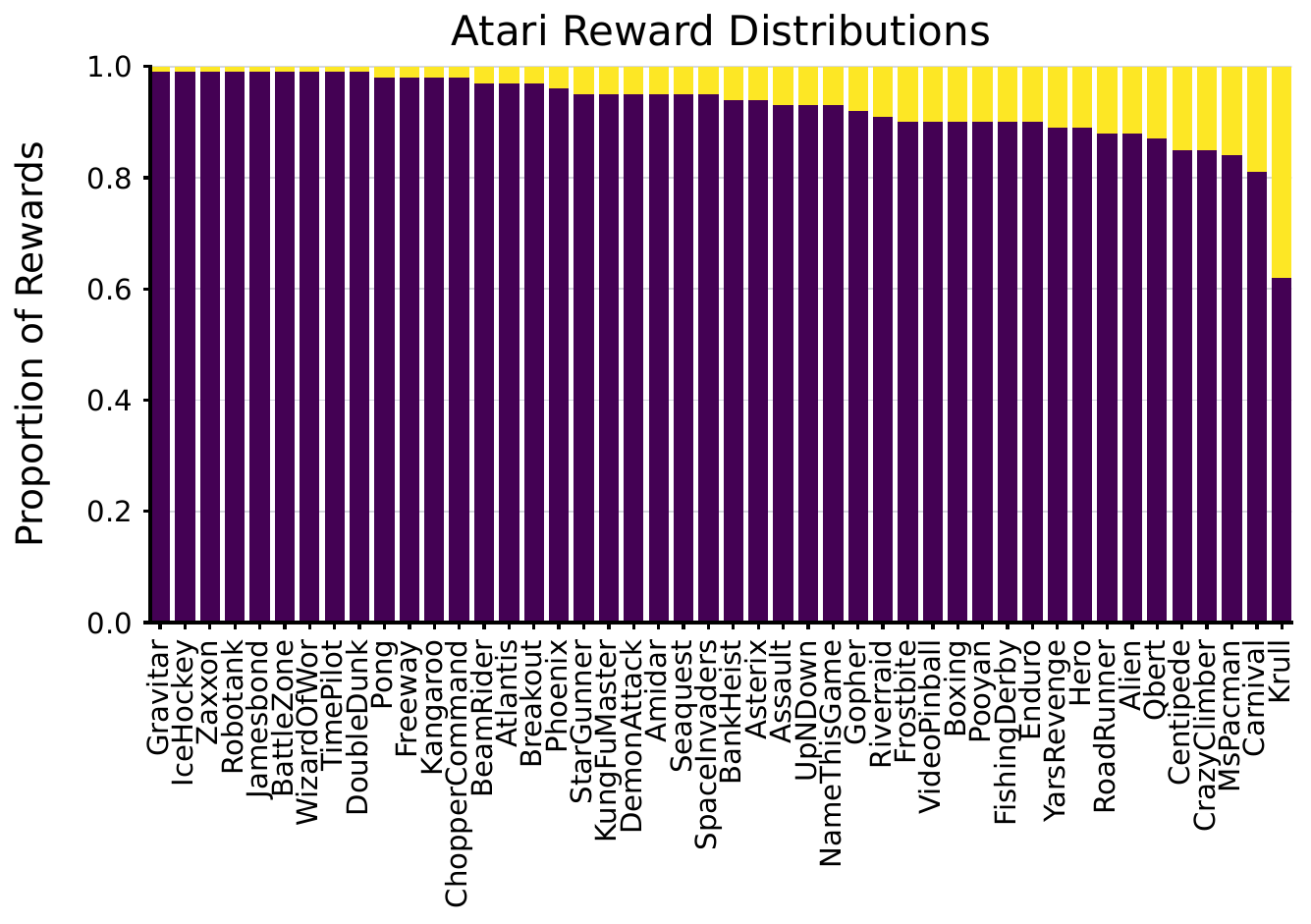}
    \end{minipage}
    \hfill
    \begin{minipage}{0.47\linewidth}
    \vspace{-0.235cm}
    \includegraphics[width=\textwidth]{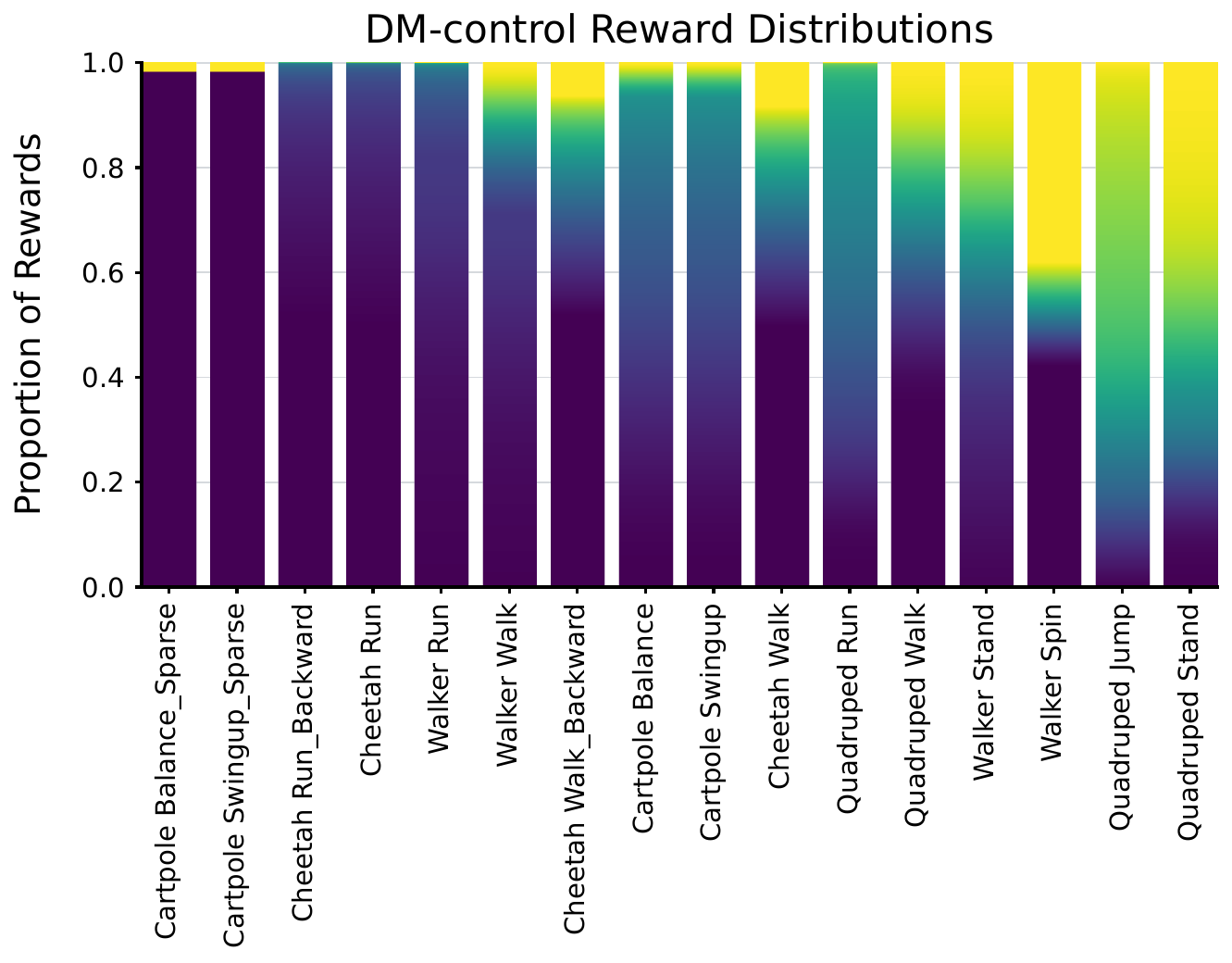}
    \end{minipage}
    \caption{Comparison of reward distributions between ALE (left) and \rebuttal{the DM-control (right)}. For each environment $1$M rewards are collected from exploratory agents. Each color in the plot corresponds to a reward value between $0$ and $1$ with the height of that color corresponding to the relative proportion of that reward in the dataset, i.e., the quantile function of the empirical reward distribution.}\label{fig:atari-reward-dist}
\end{figure}

Additionally, the reward structures in these environments tend to be much denser than in the ALE. In \cref{fig:atari-reward-dist} we plot the reward distributions for an exploratory policy in both the Arcade Learning Environment \citep{bellemare13ale} and the DeepMind Control Suite \citep[DMC;][]{tunyasuvunakool20dmc}. Specifically, for the ALE we take the rewards collected in the first 1M frames for all games in the RL Unplugged dataset \citep{gulcehre20rlu} corresponding to the exploratory phase of a DQN agent. For DMC we leverage the ExoRL dataset \citep{yarats22exorl} and collect rewards on the Cheetah, Walker, Quadruped\rebuttal{, and Cartpole} domains from an exploratory random network distillation policy. Figure~\ref{fig:atari-reward-dist} shows that the proportion of rewards that are $0$ in Atari is higher than in \rebuttal{most of the DMC tasks}, indicating that rewards are relatively more sparse in Atari.

In addition to robotics/locomotion tasks, there have been a number of recent environments simulating real-world continuous control scenarios. These include optimal control problems (continuous in both time and space) \citep{howe22myriad,ma2024transformer}, simulated industrial manufacturing and process control \citep{zhang2022smpl}, power consumption optimization \citep{moriyama18reinforcement}, process control \citep{bloor24pcgym}, dynamic algorithm configuration \citep{eimer21dacbench}, among others.

\section{Research directions}
\label{sec:researchDirections}

Since its release, the Arcade Learning Environment has been extensively used by the research community to explore fundamental problems in decision making.
However, most of this research has focused specifically on value-based methods with discrete action spaces.
On the other hand, many of the challenges presented by the ALE, such as exploration and representation learning, are not always central to existing continuous control benchmarks (see discussion in \cref{sec:comparisonCCEnvs}). In this section, we identify several research directions that the CALE facilitates. While many of these questions can be explored in different environments, the advantage of the CALE is that it has a {\em direct} analogue in the ALE, thereby enabling a more direct comparison of continuous- and discrete-control methods.

\paragraph{Exploration} As discussed in \cref{sec:exploration}, $\epsilon$-greedy is the default exploration strategy used by discrete-action agents on the ALE. Despite the existence of a number of more sophisticated methods, \citet{taiga2020bonus} argues that these were over-fit to well-known hard exploration games such as Montezuma's Revenge; they demonstrated that, when aggregating with easier exploration games, $\epsilon$-greedy out-performs the more sophisticated methods. In contrast, the results in \cref{sec:exploration} demonstrate that $\epsilon$-greedy under-performs simply sampling from $\mu$ in SAC. This may be an instance of {\em policy churn}, which has been shown to have implicit exploratory benefits in discrete-action agents \citep{schaul22churn}. Interestingly, our results show that for SAC-D (SAC with discrete actions explored in \cref{sec:comparisonToDiscrete}), $\epsilon$-greedy outperforms sampling from the categorical distribution for exploration (see \cref{sec:sacdResults}). We believe the CALE provides a novel opportunity for developing exploration methods for continuous-control agents in non-robotics tasks.

\paragraph{Network architectures} 
Recent work has demonstrated the value in exploring alternative network architectures for RL agents \citep{espeholt18impala,graesser22state,obando2024prunedRL,obando2024mixtures}. Similarly, notions of ``learned representations'' \citep{schwarzer2020data,castro2021mico,zhang2021learning,yarats21improving,farebrother2023protovalue} may benefit from different techniques based on the type of action space and losses used (a fact confirmed by the results in \cref{fig:encoderComparison}). Indeed, \citet{farebrother2024stop} demonstrated a stark performance difference resulting from switching from regression to classification losses; given their evaluations was limited to value-based discrete-action agents, it remains an open question whether similar findings carry over to continuous action spaces.

\paragraph{Offline RL} Offline RL, where RL agents trained on a fixed dataset \citep{levine2020offline}, has seen a significant growth in interest over the last few years. One of the main challenges in this setting is when there is insufficient state-action coverage in the dataset; this is particularly pronounced in discrete-action settings, where there is no clear notion of {\em similarity} between actions. Continuous control settings perhaps do provide a more immediate notion of action similarity, which could help mitigate out-of-distribution issues in offline RL. For instance, would the tandem effect \citep{ostrovski2021difficulty} still be present when training from offline data in continuous action spaces?

\paragraph{Plasticity} \citet{nikishin22primacy} demonstrated that SAC benefits from full network resets in MuJoCo, where a multi-layer perceptron network is used. For SPR \citep{schwarzer2020data} on the 100k ALE, the authors originally had to limit resets to the penultimate (dense) layer; only by switching to shrink and perturb \citep{ash20warm} does this network benefit from ``resets'' \citep{doro2023sampleefficient,schwarzer23bbf}. An interesting question is whether the benefit of full resets are tied to the use of a continuous-control actor-critic method like SAC, or to the fact that only dense layers are needed for MuJoCo. More generally, do findings related to plasticity loss \citep{sokar23dormant,lyle23understanding} apply equally to discrete- and continuous-control agents?

\paragraph{Action parameterizations}
The choice of Gaussian distributions for each of the action dimensions, initialized in the middle of the action ranges (as used by SAC) is by no means the only option.
It would be interesting to explore alternative action parameterizations\rebuttal{, different inductive biases, and evaluate agents already making use of similar re-parameterizations \citep{hafner2020Dream}.}

\section{Discussion}
\label{sec:discussion}
Academic benchmarks in machine learning are meant to provide a standardized and reproducible methodology with which to evaluate and compare algorithmic advances. In RL, these benchmarks have historically been divided between those suitable for discrete control (such as the ALE), and those suitable for continuous control (such as MuJoCo and DM-Control)\footnote{\rebuttal{\citet{tang2020discretizing} showed that discretizing actions can improve performance on DMC tasks.}}. This has made it difficult to directly compare the performance of these two types of algorithms, resulting in less transfer of advances between the continuous- and discrete-control communities than one would hope for.

One of the advantages of the CALE is that it provides a {\em unified} suite of environments on which to evaluate both types of algorithms, given that both the ALE and the CALE use the same underlying joystick events and Stella emulator. The ALE has been used in a large number of research papers, and there is a growing sentiment that it is no longer interesting; the CALE provides a fresh take on this benchmark, while benefiting from the familiarity that the community already has with it.

One could argue that human evaluations, introduced by \citet{mnih2015humanlevel} and used to normalize most ALE experiment scores, are more relevant with the CALE since the human evaluator presumably played on a real joystick. Given that our SAC baseline achieves only 0.4 IQM (where a 1.0 indicates human-level performance), the CALE provides a new challenge to achieve human-level performance on the suite of Atari 2600 games, and aid in the development of generally capable agents.

\paragraph{Limitations} One limitation of this work is the number of baselines evaluated. We used the Dopamine framework \citep{castro18dopamine} for our evaluations, which unfortunately only provides SAC and a recently added implementation of PPO as continuous-control agents. It would be useful to evaluate other continuous-control agents, as well as other agent implementations, on the CALE to build a broader set of baselines for future research. While most games use the joystick illustrated in \cref{fig:actionMappings}, Pong and Breakout were originally played on non-discrete {\em paddles} \citep{montfort09racing}; for this version of the CALE we decided to maintain the same action dynamics across all games, but it would be interesting to add support for paddles, where continuous actions are no longer mapped to discrete events.\footnote{In the ALE, discrete actions are mapped to hard-coded paddle displacements, which we replicated in our implementation.}

\newpage

\paragraph{Acknowledgements}
The authors would like to thank Georg Ostrovski for providing us with a valuable review of an initial version of this work. Additionally, we thank Hugo Larochelle, Marc G. Bellemare, Harley Wiltzer, Doina Precup, and the Google DeepMind Montreal team for helpful discussions during the preparation of this submission. We would also like to thank the Python community \citep{van1995python, oliphant07python} for developing tools that enabled this work, including NumPy \citep{harris2020array}, Matplotlib \citep{hunter2007matplotlib} and JAX \citep{bradbury2018jax}. \rebuttal{Finally, we would like to thank Mark Towers and Jet and the Farama Foundation for their help reviewing the code to integrate CALE into the ALE.}

\bibliographystyle{plainnat}
\bibliography{cale}

\newpage
\appendix

\section{How to run CALE}
\label{sec:howto}

CALE is included as of version 0.10 of the Arcade Learning Environment \citep{bellemare13ale} which can be installed with the command 
\verb|pip install ale-py|. 
A Gymnasium \citep{towers23gymnasium} interface is also provided and can be installed via 
\verb|pip install gymnasium[atari]|.
Once installed the keyword argument \verb|continuous| can enable continuous actions as shown in the code below. 


\lstset{language=Python, style=prettypython}
\begin{lstlisting}[frame=single]
import gymnasium

# `env.action_space` will be continuous
env = gymnasium.make("Pong-v5", continuous=True)

\end{lstlisting}

\section{Code specifications}
\label{sec:codeSpecifications}
The implementation of CALE is available \rebuttal{as part of the ALE: \url{https://github.com/Farama-Foundation/Arcade-Learning-Environment} (under GPL-2.0 license)}.

For SAC and PPO, we used the Dopamine \citep{castro18dopamine} implementations. Taking Dopamine's root directory \url{https://github.com/google/dopamine/}, the specific code paths used are:
\begin{itemize}
    \item The SAC implementation is available at \href{https://github.com/google/dopamine/blob/master/dopamine/labs/cale/sac_cale.py}{labs/cale/sac\_cale.py} 
    \item The PPO implementation is available at \href{https://github.com/google/dopamine/blob/master/dopamine/labs/cale/ppo_cale.py}{labs/cale/ppo\_cale.py} 
    \item All networks used are available at \href{https://github.com/google/dopamine/blob/master/dopamine/labs/cale/networks.py}{labs/cale/networks.py}
    \item For SAC-D we simply modified the SAC actor outputs to emit a categorical distribution with \verb|jax.random.categorical|. From this, we can easily extract the log probabilities with \verb|jax.nn.log_softmax|, and select actions greedily with \verb|jnp.argmax|.
\end{itemize}

\section{Hyper-parameters}
\label{sec:sacHypers}

In the following table we specify the hyper-parameters used for the various agents considered. For the most part we used the default hyper-parameters specified in the Dopamine gin files for \href{https://github.com/google/dopamine/blob/master/dopamine/labs/atari_100k/configs/DER.gin}{DER}, \href{https://github.com/google/dopamine/blob/master/dopamine/jax/agents/dqn/configs/dqn.gin}{DQN}, and \href{https://github.com/google/dopamine/blob/master/dopamine/jax/agents/sac/configs/sac.gin}{SAC}. For SAC-D, we modified settings according to what was suggested by \citet{christodoulou19sac}. 

The full hyper-parameter specifications for SAC are available at \href{https://github.com/google/dopamine/blob/master/dopamine/labs/cale/configs/sac_cale.gin}{labs/cale/configs/sac\_cale.gin} and \href{https://github.com/google/dopamine/blob/master/dopamine/labs/cale/configs/sac_cale_100k.gin}{labs/cale/configs/sac\_cale\_100k.gin}.

The full hyper-parameter specifications for PPO are available at \href{https://github.com/google/dopamine/blob/master/dopamine/labs/cale/configs/ppo_cale.gin}{labs/cale/configs/ppo\_cale.gin} and \href{https://github.com/google/dopamine/blob/master/dopamine/labs/cale/configs/ppo_cale_100k.gin}{labs/cale/configs/ppo\_cale\_100k.gin}.

\begin{table}[!h]
 \centering
  \caption{Hyper-parameter setting for all agents.}
  \label{tbl:defaultvalues}
 \begin{tabular}{@{} cccccc @{}}
    \toprule
  Hyper-parameter &  DER & DQN & SAC & SAC-D & PPO \\
    \midrule
     Adam $\epsilon$ & 0.00015 & 1.5e-4 &  1.5e-4 & 1.5e-4 & 1e-5 \\
     Batch Size & 32 & 32 & 32 & 64 & 1024 \\
     Number of hidden units & 512 & 512 & 512 & 512 & 512 \\
     Discount Factor & 0.99 & 0.99 & 0.99 & 0.99 & 0.99 \\
     Learning Rate & 0.0001 & 6.25e-5 & 6.25e-5 & 0.0003 & 2.5e-4 \\
     Exploration $\epsilon$ & 0.01 & 0.01 & 0.01 & 0.01 & 0.01 \\
     Minimum Replay History & 1600 & 20000 & 20000 & 20000 & - \\
     Update Horizon & 10 & 1 & 1 & 1 & - \\
     Update Period & 1& 4 & 4 & 4 & - \\
     \bottomrule
  \end{tabular}
\end{table}

\newpage
\section{ALE game specifications}
\label{sec:aleGameSpecs}

In the following list we indicate the minimum action values for each game. Games with an asterisk next to them are games which are part of the 26 games for the Atari 100K benchmark \citep{kaiser2020model}.

\begin{itemize}
    \item AirRaid (6)
    \item Alien* (18)
    \item Amidar* (10)
    \item Assault* (7)
    \item Asterix* (9)
    \item Asteroids (14)
    \item Atlantis (4)
    \item BankHeist* (18)
    \item BattleZone* (18)
    \item BeamRider (9)
    \item Berzerk (18)
    \item Bowling (6)
    \item Boxing* (18)
    \item Breakout* (4)
    \item Carnival (6)
    \item Centipede (18)
    \item ChopperCommand* (18)
    \item CrazyClimber* (9)
    \item DemonAttack* (6)
    \item DoubleDunk (18)
    \item ElevatorAction (18)
    \item Enduro (9)
    \item FishingDerby (18)
    \item Freeway* (3)
    \item Frostbite* (18)
    \item Gopher* (8)
    \item Gravitar (18)
    \item Hero* (18)
    \item IceHockey (18)
    \item Jamesbond* (18)
    \item JourneyEscape (16)
    \item Kangaroo* (18)
    \item Krull* (18)
    \item KungFuMaster* (14)
    \item MontezumaRevenge (18)
    \item MsPacman* (9)
    \item NameThisGame (6)
    \item Phoenix (8)
    \item Pitfall (18)
    \item Pong* (6)
    \item Pooyan (6)
    \item PrivateEye* (18)
    \item Qbert* (6)
    \item Riverraid (18)
    \item RoadRunner* (18)
    \item Robotank (18)
    \item Seaquest* (18)
    \item Skiing (3)
    \item Solaris (18)
    \item SpaceInvaders (6)
    \item StarGunner (18)
    \item Tennis (18)
    \item TimePilot (10)
    \item Tutankham (8)
    \item UpNDown* (6)
    \item Venture (18)
    \item VideoPinball (9)
    \item WizardOfWor (10)
    \item YarsRevenge (18)
    \item Zaxxon (18)
\end{itemize}

\newpage
\section{Per-game results}
\label{sec:perGameResults}

\begin{figure}[!h]
    \centering
    {\includegraphics[width=\textwidth]{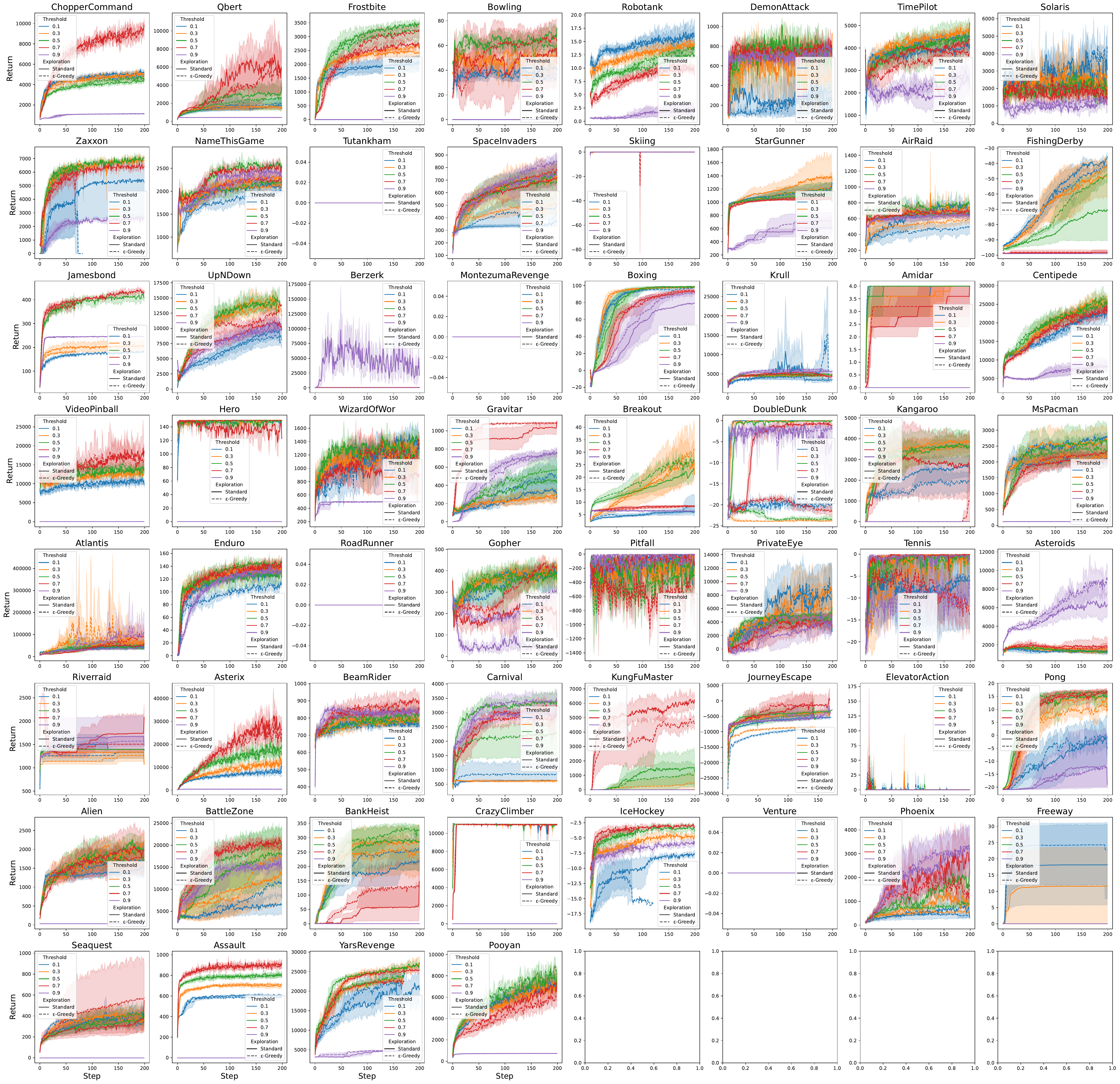}}
    \caption{Per-game learning curves for agents trained on 200M.}
    \label{fig:allGames200m}
\end{figure}

\newpage
\section{SAC-D extra results}
\label{sec:sacdResults}

\begin{figure}[!h]
    \centering
    {\includegraphics[width=0.9\textwidth]{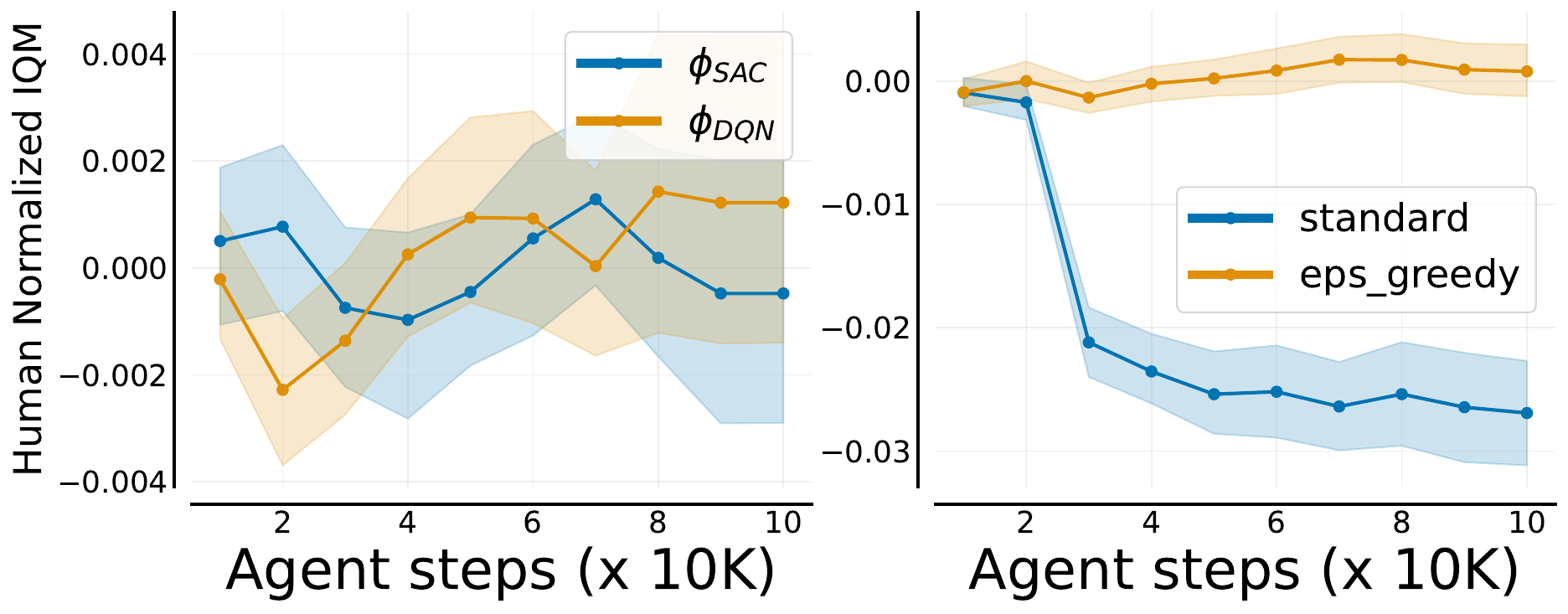}}
    \caption{{\bf Left:} Comparison of encoders on SAC-D with $\epsilon$-greedy exploration; {\bf Right:} Comparison of exploration strategies with the $\psi_{DQN}$ encoder. Reporting IQM averaged over the 26 Atari 100K games 5 runs with 95\% stratified bootstrap intervals \citep{agarwal21deep}.}
    \label{fig:sacD100kComparison}
\end{figure}

\newpage

\section*{Checklist}


\begin{enumerate}

\item For all authors...
\begin{enumerate}
  \item Do the main claims made in the abstract and introduction accurately reflect the paper's contributions and scope?
    \answerYes{}
  \item Did you describe the limitations of your work?
    \answerYes{}
  \item Did you discuss any potential negative societal impacts of your work?
    \answerYes{}
  \item Have you read the ethics review guidelines and ensured that your paper conforms to them?
    \answerYes{}
\end{enumerate}

\item If you are including theoretical results...
\begin{enumerate}
  \item Did you state the full set of assumptions of all theoretical results?
    \answerNA{}
	\item Did you include complete proofs of all theoretical results?
    \answerNA{}
\end{enumerate}

\item If you ran experiments (e.g. for benchmarks)...
\begin{enumerate}
  \item Did you include the code, data, and instructions needed to reproduce the main experimental results (either in the supplemental material or as a URL)?
    \answerYes{}
  \item Did you specify all the training details (e.g., data splits, hyperparameters, how they were chosen)?
    \answerYes{}
	\item Did you report error bars (e.g., with respect to the random seed after running experiments multiple times)?
    \answerYes{}
	\item Did you include the total amount of compute and the type of resources used (e.g., type of GPUs, internal cluster, or cloud provider)?
    \answerYes{}
\end{enumerate}

\item If you are using existing assets (e.g., code, data, models) or curating/releasing new assets...
\begin{enumerate}
  \item If your work uses existing assets, did you cite the creators?
    \answerYes{}
  \item Did you mention the license of the assets?
    \answerYes{}
  \item Did you include any new assets either in the supplemental material or as a URL?
    \answerYes{}
  \item Did you discuss whether and how consent was obtained from people whose data you're using/curating?
    \answerNA{}
  \item Did you discuss whether the data you are using/curating contains personally identifiable information or offensive content?
    \answerNA{}
\end{enumerate}

\item If you used crowdsourcing or conducted research with human subjects...
\begin{enumerate}
  \item Did you include the full text of instructions given to participants and screenshots, if applicable?
    \answerNA{}
  \item Did you describe any potential participant risks, with links to Institutional Review Board (IRB) approvals, if applicable?
    \answerNA{}
  \item Did you include the estimated hourly wage paid to participants and the total amount spent on participant compensation?
    \answerNA{}
\end{enumerate}

\end{enumerate}

\paragraph{Impact statement} This paper presents a new benchmark to promote the advancement of the field of Reinforcement Learning, and Machine Learning in general. There are many potential societal consequences of our work, none which we feel must be specifically highlighted here.

\end{document}